%% file: main.tex
\documentclass[letterpaper]{article} 
\usepackage{aaai2026}  
\usepackage{times}  
\usepackage{helvet}  
\usepackage{courier}  
\usepackage[hyphens]{url}  
\usepackage{graphicx} 
\usepackage{natbib}  
\usepackage{caption} 
\usepackage{booktabs}
\usepackage{amsmath}
\usepackage{amssymb}
\usepackage{algorithm}
\usepackage{algorithmic}
\usepackage{makecell}    
\usepackage{xcolor}     

\usepackage{newfloat}
\usepackage{listings}
\DeclareCaptionStyle{ruled}{labelfont=normalfont,labelsep=colon,strut=off} 
\floatstyle{ruled}
\newfloat{listing}{tb}{lst}{}
\floatname{listing}{Listing}

\lstdefinestyle{mypython}{
    language=Python,
    basicstyle=\ttfamily\normalsize,
    keywordstyle=\color{blue},
    stringstyle=\color{red},
    commentstyle=\color{green!70!black},
    backgroundcolor=\color{gray!5},
    showstringspaces=false,
    captionpos=b
}
\nocopyright 

\newcommand{\myattn}{EG-MLA}

\title{EG-MLA: Embedding-Gated Multi-head Latent Attention for Scalable and Efficient LLMs}

\author {
    Zhengge Cai\textsuperscript{\rm 1,\rm 2}
    Haowen Hou\textsuperscript{\rm 1},
}
\affiliations{
    \textsuperscript{\rm 1}Guangdong Laboratory of Artificial Intelligence and Digital Economy (SZ)\\
    \textsuperscript{\rm 2}Shenzhen University\\
}

\usepackage{bibentry}

\begin{document}

\maketitle

\begin{abstract}
Reducing the key-value (KV) cache size is a crucial step toward enabling efficient inference in large language models (LLMs), especially under latency and memory constraints. While Multi-Head Attention (MHA) offers strong representational power, it incurs significant memory overhead. Recent work on Multi-head Latent Attention (MLA) mitigates this by compressing KV representations into a shared latent space, achieving a better trade-off between performance and cache efficiency.
While MLA already achieves significant KV cache reduction, the scope for further compression remains limited without performance loss.
In this paper, we propose \textbf{Embedding-Gated Multi-head Latent Attention (EG-MLA)}, a novel extension of MLA that further reduces KV cache size while enhancing representational expressiveness. EG-MLA introduces a token-specific embedding gating mechanism applied in the latent space, enabling fine-grained modulation of compressed KV vectors with minimal additional computation.
Compared to MHA, EG-MLA achieves over 91.6\% reduction in KV cache size with negligible performance degradation. Relative to MLA, EG-MLA consistently improves task accuracy across diverse reasoning benchmarks while achieving up to 59.9\% additional memory savings. Our theoretical analysis highlights how embedding gating induces implicit high-order interactions, and empirical evaluations demonstrate robust generalization across model scales and compression regimes.
Notably, we successfully scale EG-MLA to over 1 billion parameters, demonstrating its practical viability for large-scale LLM deployment.
These results establish EG-MLA as a memory- and compute-efficient attention mechanism that enables scalable, high-performance inference in modern LLMs.
\end{abstract}



\input{1_introduction}

\input{2_related_work}

\input{3_method}

\input{4_experiment}

\input{5_efficiency}

\input{6_conclusion}

\bibliography{main}

\input{7_appendix}

\end{document}

%% file: 1_introduction.tex
\section{Introduction}

\begin{figure*}[thb]
    \centering
    \includegraphics[width=\linewidth]{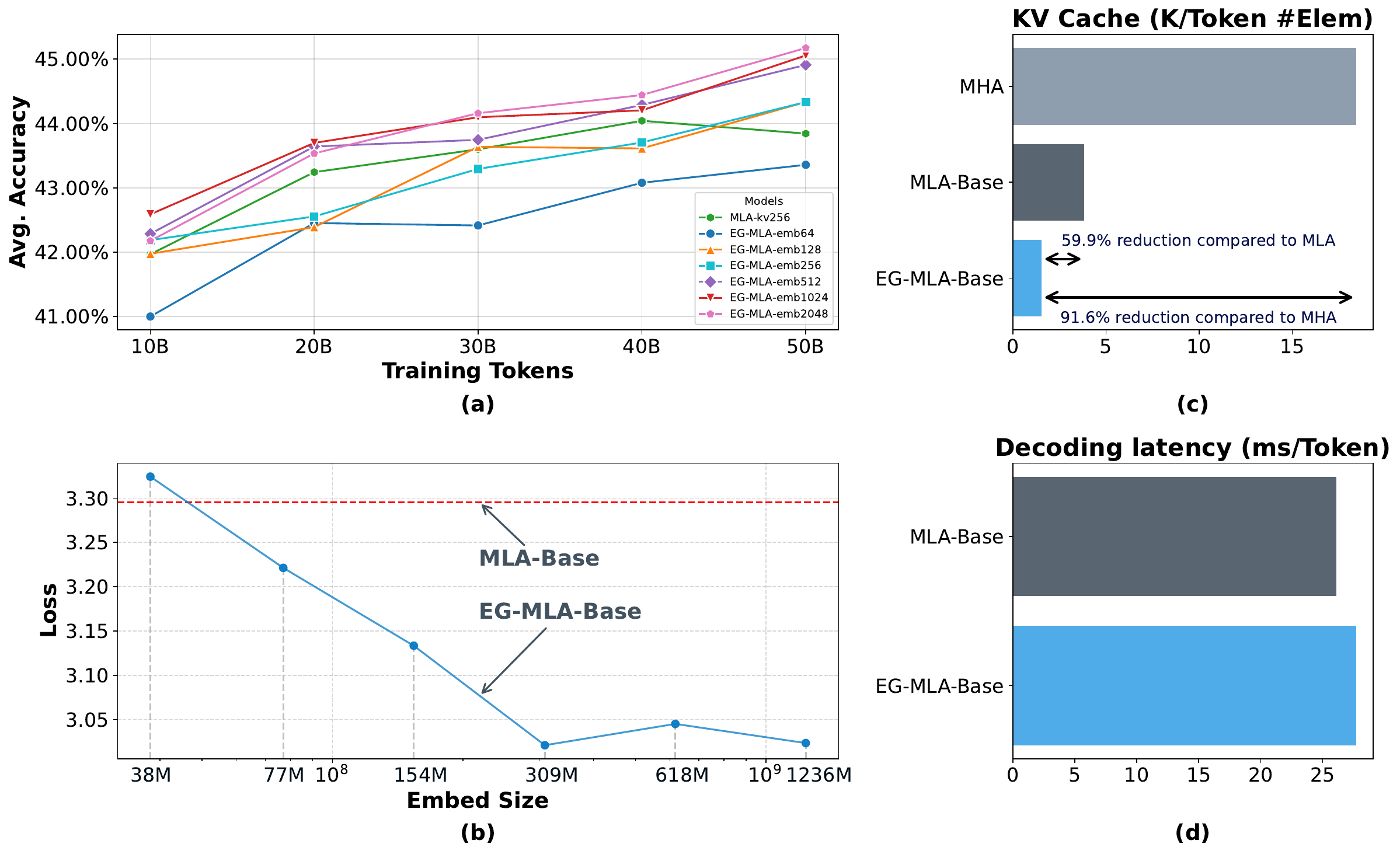}
    \caption{
(a) Average accuracy versus training tokens.
(b) Validation loss with respect to embedding size used in embedding gating. 
(c) KV cache elements per token. EG-MLA significantly reduces cache size compared to MHA and MLA. 
(d) Decoding efficiency of EG-MLA.
}
    \label{fig:teaser_plot}
\end{figure*}

Autoregressive decoding remains a critical bottleneck in the deployment of Transformer-based large language models (LLMs), particularly in real-time or resource-constrained scenarios. A major contributor to this inefficiency is the large memory footprint required to store key-value (KV) pairs during inference. In standard \textbf{Multi-Head Attention} (MHA)~\cite{vaswani2017attention}, each attention head maintains its own set of key and value vectors, leading to quadratic memory scaling with respect to sequence length and model size. While effective in capturing complex dependencies, this design is increasingly impractical for latency-sensitive applications or deployment on edge devices.

To mitigate this challenge, a series of efficient attention mechanisms have been proposed to reduce KV cache requirements while preserving model performance. Notably, \textbf{Multi-Query Attention} (MQA)~\cite{shazeer2019fast} and \textbf{Grouped-Query Attention} (GQA)~\cite{ainslie2023gqa} compress the KV cache by sharing key and value representations across attention heads or head groups. More recently, \textbf{Multi-head Latent Attention} (MLA)~\cite{liu2024deepseek} introduces a shared latent projection space that significantly reduces memory usage while maintaining strong expressiveness, striking a better balance between performance and efficiency.

Building on this line of work, we introduce \textbf{Embedding-Gated Multi-head Latent Attention (EG-MLA)}, a novel extension to MLA that further enhances memory efficiency without sacrificing model quality. EG-MLA augments the latent attention mechanism with a token-specific \emph{embedding gating module} that modulates compressed KV representations in a fine-grained manner. This lightweight gating design introduces minimal additional computation and parameter overhead, yet substantially improves the expressive capacity of the compressed latent space.

Compared to MHA, EG-MLA achieves over \textbf{91.6\% reduction in KV cache size}, and compared to MLA, it provides up to \textbf{59.9\% additional memory savings} while achieving equal or better performance across multiple reasoning benchmarks. Our theoretical analysis reveals that embedding gating implicitly induces a second-order feature space, which enhances token-specific representation capacity. Empirically, we demonstrate that EG-MLA is robust across varying compression ratios, embedding dimensions, and model scales.

Notably, we successfully scale EG-MLA to over \textbf{1 billion parameters}, confirming its effectiveness not only in compact models but also in large-scale LLMs. Despite aggressive compression, our 1.2B EG-MLA model maintains competitive performance with its MLA counterpart, while using only 40\% of the KV cache.

Our contributions can be summarized as follows:
\begin{itemize}
    \item We propose \textbf{EG-MLA}, an embedding-gated extension to latent attention that enables token-specific modulation of compressed KV representations.
    \item We demonstrate that EG-MLA achieves \textbf{superior performance-to-memory trade-offs} compared to both MHA and MLA, across diverse tasks and model scales.
    \item We provide theoretical insights into the gating mechanism and show through comprehensive experiments that EG-MLA supports efficient inference at scale, including in 1B+ parameter regimes.
    \item We identify that further reducing the KV cache in MLA is inherently difficult, as its latent space is already highly compressed and empirically optimized. To overcome this limitation without degrading model performance, we introduce a novel architectural dimension---embedding gating---which adds token-specific parameters that compensate for the loss of latent capacity. This design preserves model expressiveness and inference efficiency, enabling deeper KV compression with minimal decoding overhead.
\end{itemize}

Overall, EG-MLA advances the frontier of efficient Transformer architectures, offering a promising path toward scalable, high-performance, and resource-efficient language model deployment.

%% file: 2_related_work.tex
\section{Related Works}

\subsection{KV Cache Compression in Attention Mechanisms}
Reducing the key-value (KV) cache size during inference is a fundamental challenge in scaling large language models (LLMs). Traditional multi-head attention (MHA)~\cite{vaswani2017attention} stores separate key and value vectors for each attention head, resulting in substantial memory usage. To alleviate this, multi-query attention (MQA)~\cite{shazeer2019fast} and grouped-query attention (GQA)~\cite{ainslie2023gqa} share KV representations across heads or groups, achieving cache reduction with minimal performance degradation.

Multi-head latent attention (MLA)~\cite{liu2024deepseek, ji2025towards} extends this direction by introducing shared latent projections that significantly reduce KV cache size while retaining model expressiveness. Ji et al.~\cite{ji2025towards} further adapt MLA to arbitrary Transformer-based LLMs, highlighting its applicability for broad deployment. However, MLA’s reliance on fixed latent projections may limit its adaptability, especially under aggressive compression.

Several recent studies further advance this paradigm. TransMLA~\cite{meng2025transmla} systematizes the latent attention design for efficient inference, though it does not incorporate token-level adaptability. Deng and Woodland~\cite{deng2025multi} propose a temporal latent attention mechanism, exploring compression along different axes, which is conceptually related to our gating strategy. Mehta et al.~\cite{mehta2025latent} investigate latent attention specifically for small language models, offering evidence that such approaches can maintain quality under limited resource budgets—an area in which EG-MLA also excels.

\subsection{Gating Mechanisms in Transformer Architectures}
Gating mechanisms have been widely studied to enhance Transformer expressiveness. For example, the Highway Transformer~\cite{chai2020highway} employs self-gating to modulate internal information flow, showing that adaptive gating can improve model capacity. However, these approaches typically operate over the full attention space and introduce non-negligible computational overhead.

Our proposed EG-MLA differs by introducing a lightweight embedding gating mechanism that operates in the compressed latent space. This enables token-specific modulation of KV representations with minimal computational and memory cost, effectively enhancing expressiveness without sacrificing efficiency.

\subsection{Other Attention Variants}
Other attention variants such as Performer~\cite{choromanski2020rethinking}, Linformer~\cite{wang2020linformer}, and Longformer~\cite{beltagy2020longformer} aim to improve efficiency by approximating the softmax attention kernel. However, these methods primarily target encoder-based architectures or long-context processing, and do not directly address the memory overhead from KV caching during decoder-side autoregressive inference. In contrast, EG-MLA maintains the original attention structure while focusing on minimizing the KV cache footprint during generation. Thus, we consider our comparisons with MLA, MQA, and GQA to be more directly aligned.

\subsection{Summary}
EG-MLA integrates latent compression and token-specific modulation into a unified architecture. By building on and extending prior work in both KV cache reduction and attention gating, it achieves improved trade-offs between memory usage and model quality. This positions EG-MLA as a practical and scalable solution for efficient LLM inference across a wide range of deployment scenarios.

%% file: 3_method.tex
\section{Embedding Gated MLA}

\begin{figure}[H]
    \centering
    \includegraphics[width=\linewidth]{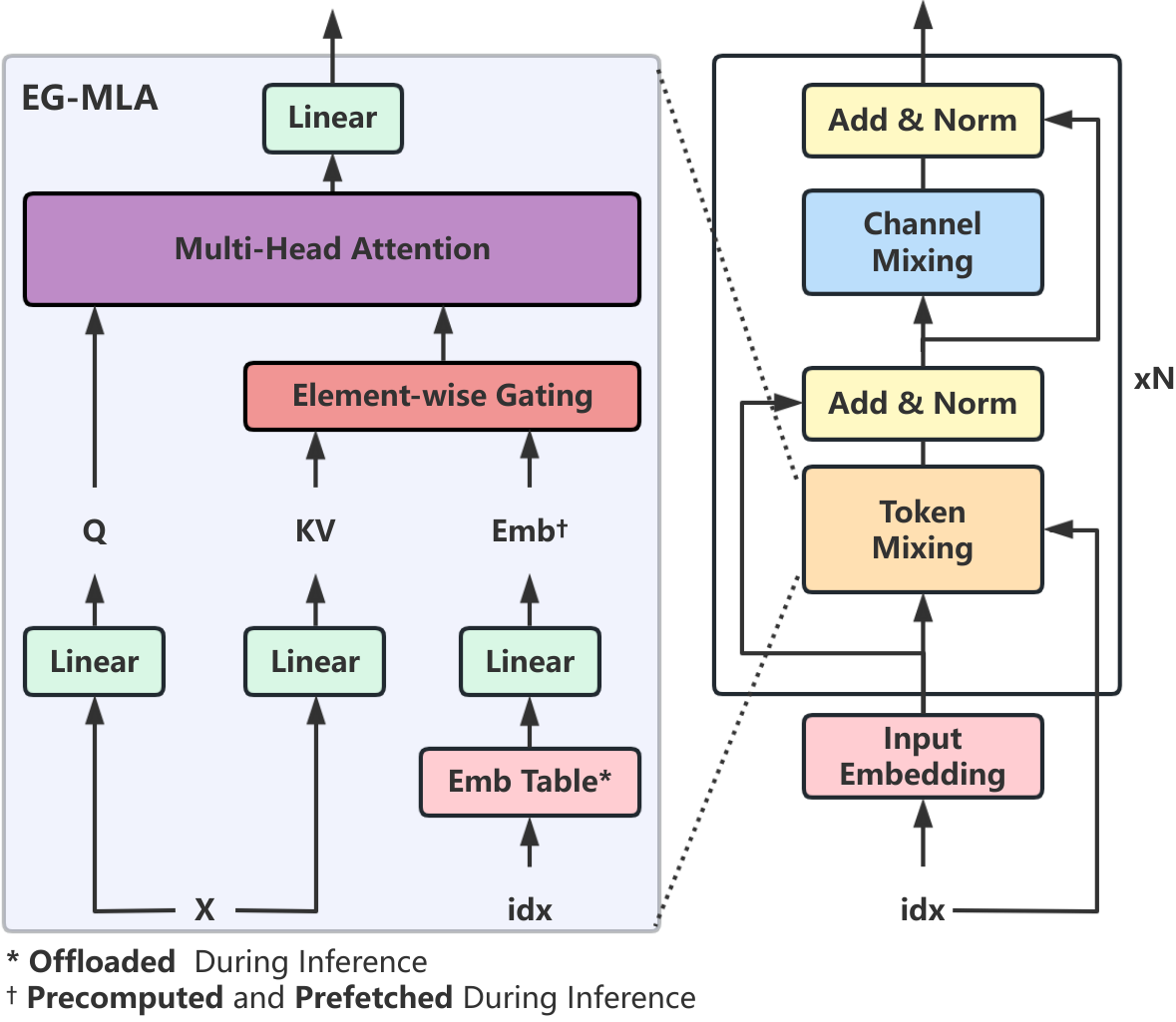}
    \caption{Simplified Architecture of the Embedding Gated Multi-Head Latent Attention (EG-MLA).}
    \label{fig:simple_eg-mla}
\end{figure}

Simplified Architecture of \myattn{} is shown in Figure~\ref{fig:simple_eg-mla}.

\subsection{Core Formulas}

\subsubsection{Low-Rank Key-Value Joint Compression} 
We begin by inheriting the core principle of MLA, namely the low-rank joint compression of keys and values, which serves to reduce the memory footprint of the KV cache:
\begin{align}
    \mathbf{c}_{t}^{KV} &= W^{DKV} \mathbf{x}_{t}, \\
    \label{eq:c_to_k}
    \mathbf{kv}_{t}^{C} &= W^{UKV} \mathbf{c}_{t}^{KV},
\end{align}
where $\mathbf{c}_{t}^{KV} \in \mathbb{R}^{d_c}$ is the compressed latent vector for keys and values; 
$W^{UKV}$ is up-projection matrix for joint keys and values.

\subsubsection{Embedding Gating}

Given the token index $i_t$, we retrieve its per-layer embedding $\mathbf{e_t}$ from a learnable lookup table.
This embedding is then passed through an up-projection matrix $W^{\text{UE}}$ to obtain a gating signal $\mathbf{g}_{t}$, representing a token-specific modulation vector.
We apply element-wise modulation to the compressed key-value representation $\mathbf{kv_t}^{C}$ via the Hadamard product with $\mathbf{g}_{t}$.
Finally, a layer normalization is applied to the result, producing the final key and value representation $\widetilde{\mathbf{kv}}_{t}^{C}$.

\begin{align}
    \mathbf{e}_{t} &= \text{Emb}(i_t), \\
    \mathbf{g}_{t} &= W^{\text{UE}} \mathbf{e}_{t}, \\
    \widetilde{\mathbf{kv}}_{t}^{C} &= \text{LN} \left( \mathbf{kv}_{t}^{C} \odot \mathbf{g}_{t} \right),
    \label{eq:emb_gate}
\end{align}

The subsequent computation follows the same procedure as MLA. To provide a comprehensive view of the \myattn{} mechanism, we present the complete set of formulas along with a detailed architectural diagram in Appendix C.

\subsection{Rationale behind Embedding Gating}
We analyze the embedding gating mechanism by reformulating its core expression, in order to highlight its ability to project inputs into a high-dimensional nonlinear feature space.

In a single layer of EG-MLA, the embedding gating operation can be written as
$(W^{\text{UKV}} \, \mathbf{c}_{t}^{\text{KV}}) \odot (W^{\text{UE}} \, \mathbf{e}_{t})$, which denotes the element-wise multiplication of two linearly projected features. 

For ease of analysis, this operation can be abstracted as \( (W_1 x_1) \odot (W_2 x_2) \), where \( x_1 \in \mathbb{R}^{d_1} \) and \( x_2 \in \mathbb{R}^{d_2} \) are two distinct input vectors, and \( W_1 \in \mathbb{R}^{d_1 \times d} \), \( W_2 \in \mathbb{R}^{d_2 \times d} \) are their corresponding projection matrices into a shared \( d \)-dimensional space.

Focusing on the case of a single output channel, the embedding gating can be further expressed as:
\begin{align}
\label{eq:eg_general}
& \quad w_1^\mathrm{T} x_1 \odot w_2^\mathrm{T} x_2 \\
\label{eq:eg_expand}
&= \left( \sum_{i=1}^{d_1} w_1^{(i)} x_1^{(i)} \right) \odot \left( \sum_{j=1}^{d_2} w_2^{(j)} x_2^{(j)} \right) \\
\label{eq:eg_sum}
&= \sum_{i=1}^{d_1} \sum_{j=1}^{d_2} w_1^{(i)} w_2^{(j)} x_1^{(i)} x_2^{(j)}
\end{align}

As shown in Eq.~\ref{eq:eg_sum}, this formulation expands into \( d_1 \times d_2 \) distinct second-order interaction terms between \( x_1 \) and \( x_2 \). Therefore, a single-layer embedding gating operation implicitly maps the input into a nonlinear feature space of dimensionality \( d_1 \times d_2 \), significantly enhancing the representational capacity without introducing any additional computational overhead. 
\textbf{This formulation forms the fundamental rationale} behind our design choice to incorporate embedding gating. By leveraging its ability to construct an \textbf{implicit high-dimensional representation}, we enable an architecture with a \textbf{more compact latent space} \textbf{without compromising expressiveness}.

%% file: 4_experiment.tex
\section{Performance Evaluation}

\subsection{Experimental Settings}
\subsubsection{Data}
We use ClimbMix~\cite{Diao2025CLIMBCI} as the training dataset, which consists of 400 billion tokens in total. For evaluation, we adopt several reasoning benchmarks: PIQA~\cite{bisk2020piqa}, ARC-C, ARC-E~\cite{Clark2018ThinkYH}, HellaSwag~\cite{Zellers2019HellaSwagCA}, WinoGrande~\cite{Sakaguchi2019WinoGrande}, and SIQA~\cite{sap-etal-2019-social}. We optimize the data mixture using the validation sets of PIQA, ARC-E, and HellaSwag, and then evaluate on their respective test sets. All evaluations are performed using the LM-Evaluation Harness~\cite{eval-harness}, under the 0-shot setting for all tasks except MMLU, which uses a 5-shot setting~\cite{allal2024smollm, cosmopediaeval2024}.

\subsubsection{Model}
We train a series of Transformer decoder-only models using the EG-MLA architecture with next-token prediction as the training objective. The training data consists of 50 billion tokens.

\subsubsection{Baselines}
We compare the proposed EG-MLA architecture against MLA models with equivalent activation parameter counts.
More detailed experimental settings and training hyper-parameters can be found in Appendix A.

\subsection{Towards Compact KV Cache}

Table~\ref{tab:scale_kv_cache_acc} reveals several noteworthy patterns regarding the impact of embedding gating and KV cache size on model performance. 
First, the introduction of embedding gating consistently enhances performance across nearly all tasks compared to the baseline MLA model with the same KV cache size, indicating that the proposed mechanism effectively enriches representational capacity without increasing computational burden. 
Second, \myattn{} models demonstrate strong robustness to aggressive KV compression: even when the KV cache size is reduced from 256 to as low as 64 or 16, the average accuracy remains stable, with only marginal fluctuations across most benchmarks. 
This suggests that embedding gating allows the model to retain crucial information in a more compact latent space. 

\begin{table*}[ht]
\resizebox{\linewidth}{!}{%
\begin{tabular}{lc|cccccccccccc|c}
\midrule
\textbf{Model Name} &
  \textbf{Size} &
  \textbf{piqa} &
  \textbf{arc-c} &
  \textbf{arc-e} &
  \textbf{hella} &
  \textbf{winog} &
  \textbf{siqa} &
  \textbf{mmlu} &
  \textbf{obqa} &
  \textbf{boolq} &
  \textbf{race} &
  \textbf{truth} &
  \textbf{lmb.o} &
  \textbf{Avg.} \\ \midrule
MLA-Base-kv256 & 120M & 69.48 & 26.45 & 61.36 & 34.11 & 52.33 & 85.40 & 25.12 & 21.40 & 58.10 & 30.91 & 29.82 & 31.63 & 43.84 \\ 
EG-MLA-Base-kv256 & 125M & 70.29 & 29.27 & 62.80 & 35.82 & 50.67 & 86.40 & 25.22 & 24.40 & 55.26 & 30.62 & 29.20 & 33.63 & 44.46 \\ 
EG-MLA-Base-kv128 & 122M & 71.49 & 29.18 & 62.75 & 35.57 & 53.43 & 87.80 & 26.71 & 22.20 & 47.74 & 30.24 & 29.09 & 33.48 & 44.14 \\ 
EG-MLA-Base-kv64 & 120M & 71.60 & 29.61 & 63.89 & 35.31 & 51.14 & 86.20 & 24.19 & 23.00 & 55.05 & 30.43 & 29.21 & 32.35 & 44.33 \\ 
EG-MLA-Base-kv16 & 119M & 71.16 & 28.50 & 62.08 & 34.48 & 50.75 & 84.70 & 26.21 & 21.40 & 57.13 & 31.29 & 30.58 & 30.95 & 44.10 \\ 
\bottomrule
\end{tabular}%
}
\caption{Accuracy of different models across 12 benchmark tasks under varying KV cache sizes. All models use a fixed embedding dimension of 256. "Size" denotes the number of activated parameters. Task abbreviations: LAMBADA (lmb.o), HellaSwag (hella), Winogrande (winog), TruthfulQA (truth).}
\label{tab:scale_kv_cache_acc}
\end{table*}

We also assess EG-MLA on standard language modeling datasets such as WikiText103~\cite{merity2018analysis}, LAMBADA~\cite{paperno2016lambada}, and ClimbMix~\cite{Diao2025CLIMBCI}. 
As reported in Appendix D, EG-MLA consistently outperforms MLA in perplexity or validation loss under equivalent compression settings. 
This suggests that embedding gating provides benefits beyond reasoning and applies broadly to language modeling scenarios.
These results demonstrate that EG-MLA remains robust even under aggressive compression settings. 
It is important to emphasize, however, that MLA itself is already near the limit of feasible compression.

\subsection{Scaling Law for Embedding Size}

Table~\ref{tab:scale_emb_acc} highlights several key trends as the embedding dimension increases. 
Most notably, performance consistently improves when moving from extremely small embeddings (e.g., 64) to moderate sizes (e.g., 512), suggesting that larger embedding vectors enable more expressive token-specific modulation through the gating mechanism. 
However, beyond a certain point—particularly past 1024 dimensions—the performance gains begin to saturate and even exhibit some inconsistency across tasks. 
For instance, while some benchmarks such as ARC-E and LAMBADA continue to improve steadily, others like MMLU and WinoGrande fluctuate or plateau. This indicates diminishing returns from simply scaling up the embedding size, likely due to a combination of overparameterization and limited benefit from increased modulation granularity. 
Interestingly, even relatively compact models with 128–256 embedding dimensions already capture most of the benefit, striking a favorable balance between performance and efficiency. 
These findings suggest that embedding size should be carefully tuned rather than blindly scaled, and that EG-MLA achieves strong performance even under strict parameter budgets.

Notably, performance does not increase monotonically as shown in Figure~\ref{fig:teaser_plot}. Beyond the point where the embedding size exceeds the total dimensionality of the key and value vectors, we observe saturation or even slight regressions on some tasks. This suggests that overparameterization beyond the combined key-value capacity yields limited benefits, revealing a current scaling limitation of our method. Exploring architectures that scale more gracefully with increased embedding capacity is an important direction for future work.More comprehensive scaling experiments are provided in Appendix G.

\begin{table*}[ht]
\resizebox{\linewidth}{!}{%
\begin{tabular}{lc|cccccccccccc|c}
\midrule
\textbf{Model Name} &
  \textbf{Embed Size} &
  \textbf{piqa} &
  \textbf{arc-c} &
  \textbf{arc-e} &
  \textbf{hella} &
  \textbf{winog} &
  \textbf{siqa} &
  \textbf{mmlu} &
  \textbf{obqa} &
  \textbf{boolq} &
  \textbf{race} &
  \textbf{truth} &
  \textbf{lmb.o} &
  \textbf{Avg.} \\ 
\midrule
EG-MLA-Base-emb64 & 38M & 69.37 & 27.47 & 60.82 & 33.96 & 51.22 & 83.90 & 27.73 & 21.00 & 55.17 & 28.71 & 29.81 & 31.13 & 43.36 \\ 
EG-MLA-Base-emb128 & 77M & 70.62 & 29.44 & 63.22 & 34.72 & 54.22 & 86.60 & 25.48 & 22.60 & 54.68 & 28.80 & 30.54 & 31.07 & 44.33 \\ 
EG-MLA-Base-emb256 & 154M & 71.60 & 29.61 & 63.89 & 35.31 & 51.14 & 86.20 & 24.19 & 23.00 & 55.05 & 30.43 & 29.21 & 32.35 & 44.33 \\
EG-MLA-Base-emb512 & 309M & 71.93 & 30.12 & 63.89 & 35.86 & 52.01 & 87.10 & 26.27 & 22.20 & 56.12 & 30.05 & 29.50 & 33.86 & 44.91 \\ 
EG-MLA-Base-emb1024 & 618M & 71.98 & 31.83 & 64.69 & 36.24 & 50.21 & 86.10 & 26.63 & 23.80 & 58.81 & 29.28 & 28.22 & 32.91 & 45.06 \\ 
EG-MLA-Base-emb2048 & 1236M & 72.69 & 30.89 & 66.37 & 36.44 & 54.07 & 85.40 & 25.43 & 22.80 & 53.61 & 28.90 & 30.78 & 34.70 & 45.17 \\ 
\bottomrule
\end{tabular}%
}
\caption{Accuracy results of EG-MLA models across 12 benchmark tasks with varying embedding dimensions. The KV cache dimension is fixed at 64. Task abbreviations: LAMBADA (lmb.o), HellaSwag (hella), WinoGrande (winog), TruthfulQA (truth).}
\label{tab:scale_emb_acc}
\end{table*}

\subsection{Ablation on Embedding Gating}
The ablation study in Table~\ref{tab:abalation_on_emb_gating} reveals two critical observations about the design of the embedding gating mechanism. 

\subsubsection{The Critical Role of Layer Normalization}
First, the presence of LayerNorm after the gating operation is essential. Its removal leads to a clear degradation in performance, suggesting that normalization is crucial for stabilizing the gated feature distribution and ensuring effective downstream attention computation.

\subsubsection{Effectiveness of element-wise multiplication}
Second, replacing the element-wise multiplication with simple addition results in a noticeable loss increase, although not as severe as removing LayerNorm. This indicates that the element-wise multiplicative interaction plays a more effective role in modulating key-value representations, likely because it enables fine-grained, dimension-wise control over the information flow.

Overall, these findings reinforce the importance of both architectural choices—the normalization step and multiplicative gating—in maximizing the benefits of the embedding-guided modulation.

\begin{table}[ht]
\centering
\resizebox{\columnwidth}{!}{%
\begin{tabular}{lccc}
\toprule
\textbf{Model Name} & \textbf{Size} & \textbf{Tokens} & \textbf{Loss (val)} \\
\midrule
MLA-Base-kv256 & 120M & 10B & 3.2156 \\
EG-MLA-Base-kv256 & 125M & 10B & 3.1609 \\
- remove LayerNorm & 125M & 10B & 3.2573 \\
- replace $\odot$ with addition & 125M & 10B & 3.1901 \\
\bottomrule
\end{tabular}%
}
\caption{Ablation on Embedding Gating. ”Size” refers to Activated Parameters}
\label{tab:abalation_on_emb_gating}
\end{table}

\subsection{Scaling to 1B Language Model}

To probe into the impact of scaling model size and training data, we train two models, namely MLA-1.2B and EG-MLA-1.2B (with 1.2 billion activated parameters and 600 million offloaded parameters). Both models are trained on 50 billion tokens from the ClimbMix dataset \cite{Diao2025CLIMBCI}. MLA-1.2B is built on the Deepseek framework \cite{liu2024deepseek}. EG-MLA-1.2B, which is designed for efficiency, utilizes merely 40.1\% of the cache size of MLA-1.2B while still achieving competitive performance.

\begin{table*}[ht]
\resizebox{\linewidth}{!}{%
\begin{tabular}{lcc|cccccccccccc|c}
\midrule
\textbf{Model Name} &
  \textbf{Size} &  %
  \textbf{Embed Size} &  %
  \textbf{piqa} &
  \textbf{arc-c} &
  \textbf{arc-e} &
  \textbf{hella} &
  \textbf{winog} &
  \textbf{siqa} &
  \textbf{mmlu} &
  \textbf{obqa} &
  \textbf{boolq} &
  \textbf{race} &
  \textbf{truth} &
  \textbf{lmb.o} &
  \textbf{Avg.} \\  
\midrule
MLA-1.2B & 1188M & 0M & 75.52 & 38.82 & 73.02 & 44.15 & 56.83 & 92.20 & 28.42 & 27.20 & 53.15 & 34.26 & 33.76 & 38.02 & 49.61 \\ 
EG-MLA-1.2B & 1182M & 618M & 75.41 & 38.74 & 72.85 & 44.88 & 56.83 & 91.00 & 26.92 & 27.80 & 59.45 & 33.97 & 31.27 & 40.25 & 49.95 \\ 
\bottomrule
\end{tabular}%
}
\caption{Accuracy results of additional models across 12 benchmark tasks for EG-MLA with over 1B parameters. "Size" denotes the number of activated parameters.}
\label{tab:over_1b}
\end{table*}

Table~\ref{tab:over_1b} demonstrates that EG-MLA-1.2B attains performance on par with MLA-1.2B across a wide array of benchmarks, featuring a remarkable enhancement in average accuracy. This competitive performance, in conjunction with the reduced cache size of EG-MLA-1.2B, emphasizes its capability to generalize proficiently under resource limitations.

In conclusion, EG-MLA-1.2B matches the performance of MLA-1.2B in both base and fine-tuned scenarios, while utilizing merely 40\% of the KV cache size of MLA-1.2B. These findings highlight the potential of memory- and compute-efficient architectures for the expansion of large language models, facilitating sustainable and resource-efficient AI development.

%% file: 5_efficiency.tex
\section{Efficiency Evaluation}

\subsection{Analytical Comparison of KV Cache Costs}

Table~\ref{tab:kv_cache_comp} presents a theoretical comparison of the KV cache cost per token across different attention mechanisms, including our proposed EG-MLA. 
Traditional methods such as Multi-Head Attention (MHA) and Grouped-Query Attention (GQA) store separate key and value vectors for each head or group, leading to a cache size that scales with the number of heads $n_h$ or groups $n_g$. 
Multi-Query Attention (MQA), while more efficient, sacrifices representational capacity by sharing keys and values across all heads.

In comparison, MLA introduces a shared latent vector to compress key-value representations, significantly reducing the cache size while maintaining reasonable performance. 
However, EG-MLA further improves this design by applying a token-specific embedding gating mechanism, enabling more expressive modulation of the shared latent without increasing cache overhead. 
Specifically, EG-MLA stores only a compressed latent vector of dimension $d_c / r$ per token, resulting in a leaner cache footprint than MLA. 
Despite the more aggressive compression, EG-MLA retains strong representational power thanks to the multiplicative interaction with token embeddings. 
This analytical comparison highlights EG-MLA’s advantage in offering both compact memory usage and enhanced modeling capacity.

The actual KV cache savings can be found in Appendix~F, where we provide a concrete example. 
In this case, EG-MLA achieves a \textbf{59.9\%} reduction in KV cache compared to MLA, and a substantial \textbf{91.6\%} reduction compared to MHA.

\begin{table}[!t]
\centering
\resizebox{\columnwidth}{!}{
\begin{tabular}{@{}l c c@{}}
\toprule
\textbf{Attention Mechanism} & \textbf{KV Cache / Token (\#Elem)} & \textbf{Capability} \\
\midrule
Multi-Head Attention (MHA) & $2 n_{h} d_{h} l$ & Strong \\
Grouped-Query Attention (GQA) & $2 n_{g} d_{h} l$ & Moderate \\
Multi-Query Attention (MQA) & ~~~~$2 d_{h} l$ & Weak \\
Multi-Head Latent Attention (MLA) & $(d_{c} + d_h^R) l$ & Strong \\
\midrule
EG-MLA (Ours) & $(d_{c}/r + d_h^R) l$ & Stronger \\
\bottomrule
\end{tabular}
}
\caption{
Comparison of the number of KV cache elements per token across different attention mechanisms. 
$n_{h}$ denotes the number of attention heads, 
$d_{h}$ the dimension per attention head, 
$l$ the number of layers, 
$n_{g}$ the number of groups in GQA, 
and $d_{c}$ and $d_h^R$ represent the latent space dimension and the KV cache dimension used for RoPE in MLA, respectively. 
For EG-MLA, a compressed latent vector of dimension $d_c / r$ is stored per token, where $r$ is the compression factor applied to the latent representation. 
This results in a significantly smaller KV cache while preserving strong representational capacity through embedding gating.
}
\label{tab:kv_cache_comp}
\end{table}

\begin{figure}[h]
    \centering
    \includegraphics[width=\columnwidth]{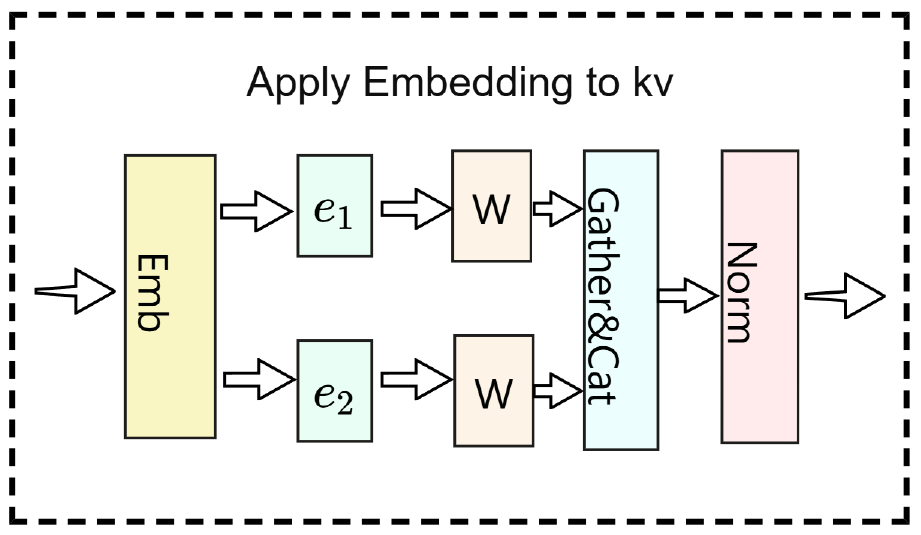}
    \caption{Dimension-wise sharding pipeline.}
    \label{fig:eg-tp}
\end{figure}

\subsection{Empirical Comparison of Efficiency}
\begin{figure*}[thb]
    \centering
    \includegraphics[width=\linewidth]{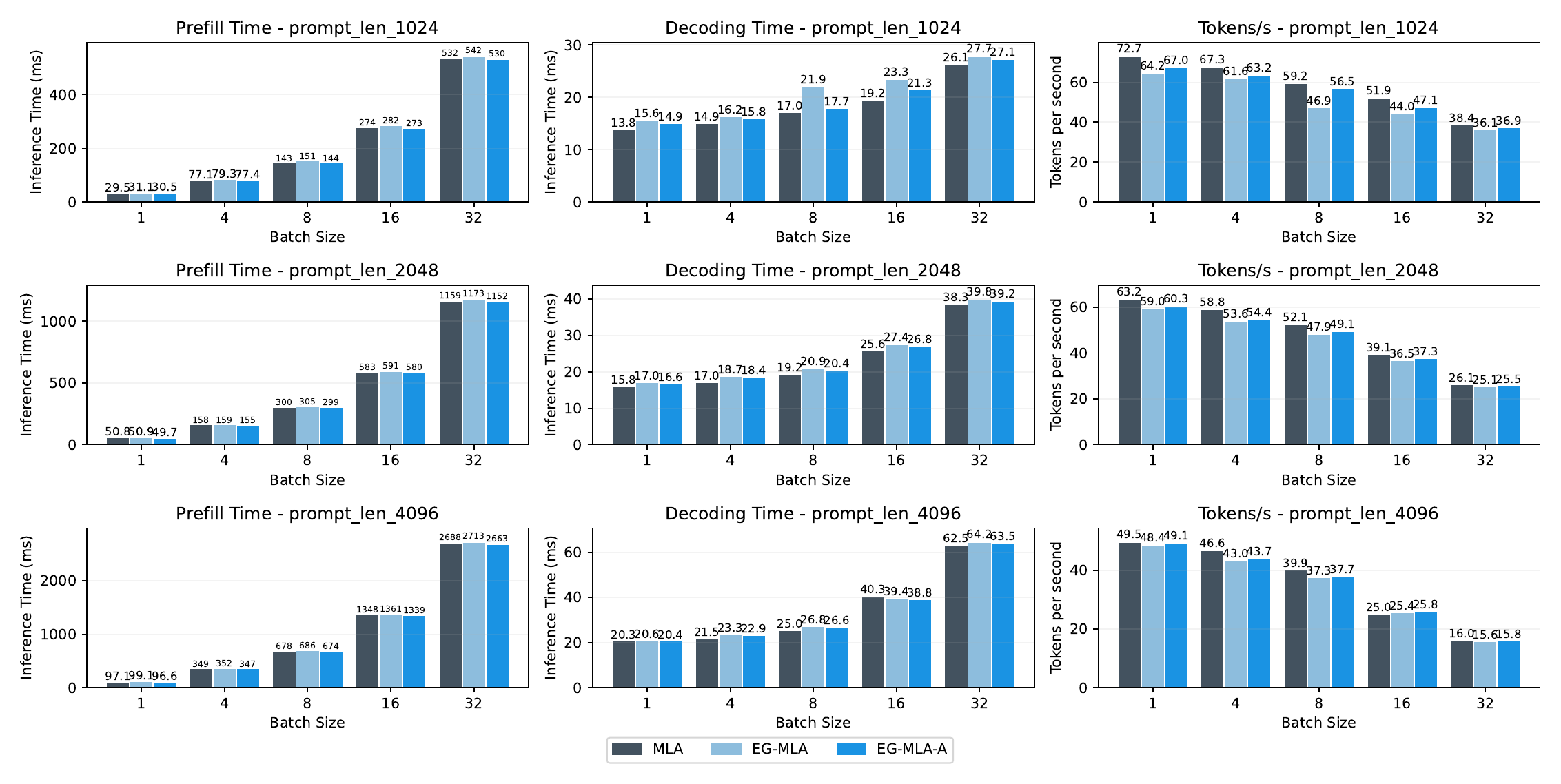}
    \caption{Prefill and Decoding times and Tokens per seconds for MLA, EG-MLA and EG-MLA-A across configurations on an NVIDIA A800 80GB GPU. EG-MLA-A is an inference-optimized variant of EG-MLA that pre-computes all embeddings at model-loading time and pre-fetch during inference.}
    \label{fig:latency_plot}
\end{figure*}

We evaluate the inference efficiency of EG-MLA on a single NVIDIA A800 80 GB GPU. As shown in Figure \ref{fig:latency_plot}, we report Prefill latency, Decoding latency and tokens-per-second under varying prompt lengths and batch sizes. The MLA model contains 120M parameters in total, whereas EG-MLA contains 274M, of which 154M are embed parameters.EG-MLA-A is an inference-optimized variant of EG-MLA in which the projection matrices are pre-computed offline during model loading; during inference, matrix multiplication is thus eliminated and only a table lookup is performed, further reducing inference latency.Despite the extra embed parameters introduced, EG-MLA maintains inference performance comparable to or better than that of MLA. Concretely, when the batch size exceeds 8, EG-MLA reduces the average pre-fill latency by 5.6 ms compared with MLA while adding only 0.91 ms to the average decoding latency. When the batch size reaches 32, EG-MLA-A narrows the average decoding latency gap with MLA to merely 1 ms, underscoring the efficiency of our approach for large-scale parallel workloads.

\begin{table}[htbp]
\centering
\resizebox{\columnwidth}{!}{
\begin{tabular}{cccccc}
\toprule
Method & Node & Peak Mem. & Throughput & Avg. Step Time & Comm. Time \\
\midrule
vanilla & 2 & 13950 & 42528 & 385.25 & -- \\
dimension-wise & 2 & 12233 & 86743 & 188.88 & 16.25 \\
vanilla & 4 & 13950 & 80114 & 409.02 & -- \\
dimension-wise & 4 & 11759 & 175031 & 187.21 & 24.38 \\
\bottomrule
\end{tabular}
}
\caption{Performance comparison between vanilla and dimension-wise parallelism under varying node counts. Memory is reported in megabytes (MB) and throughput in tokens per second (tokens/s), and all timing metrics in milliseconds (ms).}
\label{tab:tpres}
\end{table}

\subsection{Embedding Parallelism}
Embedding gating introduces additional parameters into each attention layer, significantly increasing GPU memory usage during training. To mitigate peak memory footprint on individual devices, we adopt a \emph{dimension-wise sharding} strategy~\cite{shoeybi2020megatronlmtrainingmultibillionparameter}, which partitions the key–value embedding uniformly along the embedding dimension across $W$ parallel workers. The per-rank embedding dimension is given by:
\begin{align}
    \mathbf{d}_{\text{local}} &= \frac{d_{kv}}{W},
\end{align}
where $d_{kv}$ is the original KV embedding dimension.

Each worker independently computes its local embedding and corresponding projection:
\begin{align}
    \mathbf{e}_i &= \mathrm{Emb}(i_t), \\
    \mathbf{g}_{t}^{(i)} &= W_i^{UE} \mathbf{e}_i,
\end{align}
and the final gating vector is reconstructed via concatenation.

This approach yields near-linear memory scaling. Embedding lookups and projections are fully parallelized across workers, with only a single \texttt{all-gather} operation needed during final aggregation, incurring minimal communication overhead. A simplified pipeline is illustrated in Figure~\ref{fig:eg-tp}.

As shown in Table \ref{tab:tpres}, dimension-wise parallelism with two nodes reduces peak memory by 12.3\%, shortens per-step time by 51\%, and increases throughput by 104\% relative to the baseline. Scaling to four nodes yields further reductions in both peak memory and per-step time, together with an additional throughput gain.

\paragraph{Why Dimension-wise Sharding?}
Tensor parallelism distributes matrix computations across multiple GPUs to reduce memory usage and accelerate training, but introduces inter-device communication. Compared to alternatives like layer-wise sharding, we choose dimension-wise partitioning for the following reasons:
\begin{itemize}
    \item \textbf{Load Balancing}: Evenly splits the embedding matrix, ensuring uniform computational load across devices.
    \item \textbf{No Serial Dependencies}: Enables full parallelism by avoiding sequential computation across layers.
    \item \textbf{Memory Efficiency}: Embedding layers are memory-intensive; partitioning them maximizes GPU memory bandwidth and cache utilization.
\end{itemize}

\begin{figure*}[htbp]
    \centering
    \includegraphics[width=\linewidth]{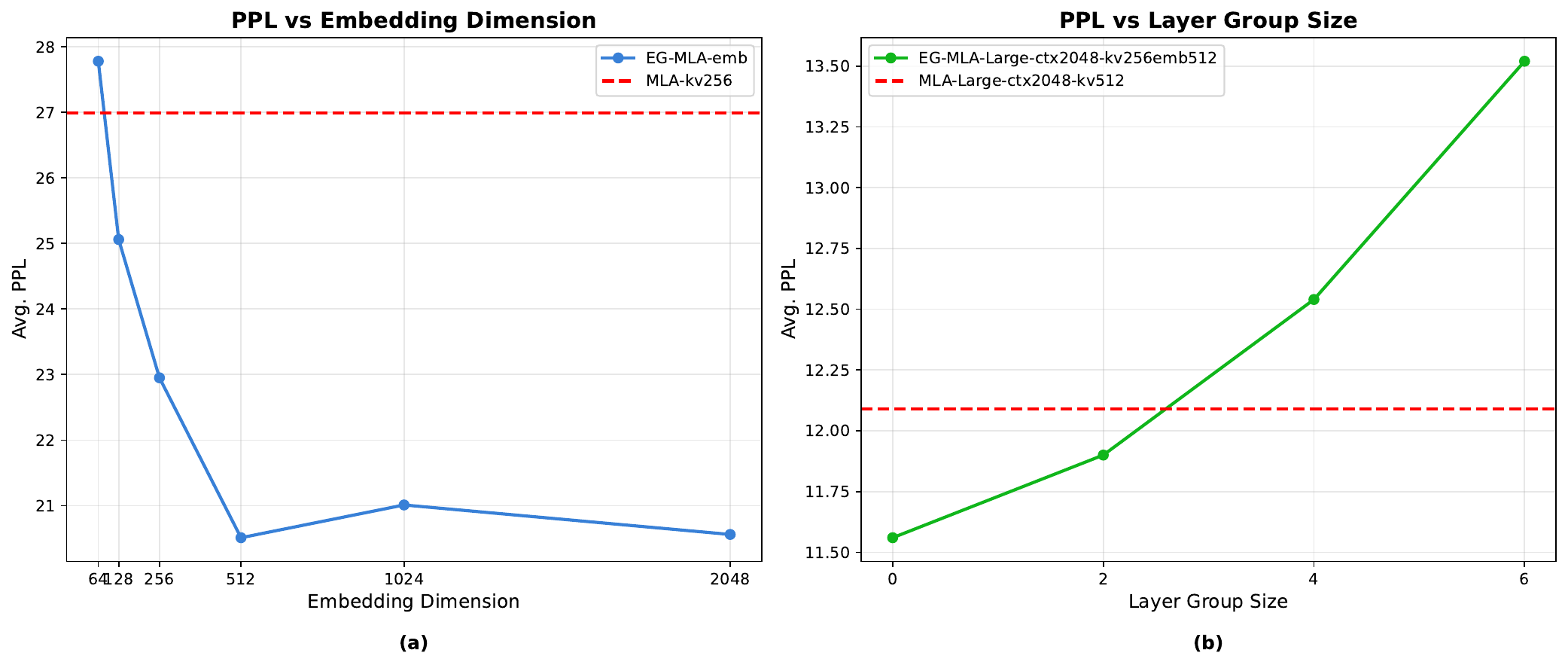}
    \caption{
(a) Average Perplexity(PPL) versus Embedding Dimension.
(b) Average Perplexity(PPL) versus Layer Group Size in Inter.
The PPL reported in both (a) and (b) are averaged over the ClimbMix-val, Wiki, and LAMBADA benchmarks.
}
    \label{fig:figure-ef}
\end{figure*}

\subsection{Toward Greater Efficiency}
To compress model parameters further while preserving performance, we explore two orthogonal directions. First, as shown in Table 5, the proposed embedding-gating mechanism yields a significantly smaller KV-cache than competing techniques. Second, we investigate inter-layer parameter sharing: the first half of the KV-projection matrices is reused across neighboring layers, with a single hyper-parameter that controls the width of the sharing window. Setting this hyper-parameter to 2, for example, theoretically halves the parameter count of the first half of the KV projections and simultaneously reuses the associated intermediate activations, cutting both memory footprint and computation. Detailed ablations are provided in Appendix H.

Figure \ref{fig:figure-ef} reveals that perplexity decreases monotonically with embedding dimension (emb) until it plateaus beyond emb = 512, indicating the current capacity limit of our embedding gating mechanism; surpassing this limit will be the focus of future work. Figure 3(b) shows that a layer group size of 2 or 3 strikes the best trade-off. Consequently, we recommend emb = 512 and layer group size = 2 for an optimal balance between performance and storage.

%% file: 6_conclusion.tex
\section{Conclusion}
In this work, we introduced \textbf{Embedding-Gated Multi-head Latent Attention (EG-MLA)}, a novel extension of MLA designed to achieve more compact and expressive key-value (KV) representations. By integrating a lightweight embedding gating mechanism, EG-MLA effectively compresses the KV cache without sacrificing model performance. Through extensive experiments across multiple reasoning benchmarks, we demonstrated that EG-MLA consistently matches or surpasses MLA while significantly reducing memory usage, achieving up to a 59.9\% reduction in KV cache compared to MLA and over 91.6\% relative to MHA. We further analyzed the impact of embedding size, layer normalization, and multiplicative gating, highlighting the critical design choices that underpin EG-MLA's effectiveness.
Although MLA already represents a compact and efficient KV caching solution, our work highlights that meaningful further compression is only achievable by innovating along other architectural axes. EG-MLA achieves this by introducing a new token-specific embedding gating dimension that improves expressiveness with minimal impact on decoding latency, thereby enabling deeper compression without performance degradation.
Scaling experiments on 1B-parameter models confirmed that EG-MLA retains competitive accuracy under stringent resource constraints, making it a promising candidate for efficient large language model deployment. Additionally, our theoretical and empirical efficiency analyses reveal that embedding gating enables a favorable trade-off between expressiveness and memory footprint. Future work will explore adaptive gating strategies, integration with sparsity-based attention mechanisms, and broader applications of EG-MLA in multi-modal and instruction-tuned models. We believe EG-MLA provides a foundation for advancing memory-efficient and high-performance language model architectures.

%% file: 7_appendix.tex
\onecolumn

\section{A. Hyper-parameters}
\label{sec:hyperparameters}

\begin{table}[ht]
\centering
\resizebox{\textwidth}{!}{%
\begin{tabular}{lccccccc}
\toprule
\textbf{Model Name} & 
\textbf{Model Size} & 
\textbf{Embed Size} & 
\makecell[c]{\textbf{n\_layer} / \\ \textbf{n\_emb} / \\ \textbf{kv\_emb\_dim}} & 
\makecell[c]{\textbf{q\_lora\_rank} / \\ \textbf{kv\_lora\_rank}} & 
\makecell[c]{\textbf{qk\_head\_dim} / \\ \textbf{v\_head\_dim}} & 
\textbf{Batch Size} & 
\textbf{Learning Rate} \\
\midrule
MLA-Base-kv256      & 120M & 0M & 12/768/0 & 0/256 & 64+64 / 64 & 0.39M & $6.0 \times 10^{-4}$ \\
MLA-Large-kv512      & 645M & 0M & 12/2048/0 & 0/512 & 128+64 / 128 & 0.39M & $2.5 \times 10^{-4}$ \\
MLA-XLarge-kv512      & 1188M & 0M & 24/2048/0 & 0/512 & 128+64 / 128 & 0.39M & $2.0 \times 10^{-4}$ \\
\midrule
EG-MLA-Base-kv256   & 279M & 154M & 12/768/256 & 0/256 & 64+64 / 64 & 0.39M & $6.0 \times 10^{-4}$ \\
EG-MLA-Base-kv128   & 276M & 154M & 12/768/256 & 0/128 & 64+64 / 64 & 0.39M & $6.0 \times 10^{-4}$ \\
EG-MLA-Base-kv64    & 274M & 154M & 12/768/256 & 0/64  & 64+64 / 64 & 0.39M & $6.0 \times 10^{-4}$ \\
EG-MLA-Base-kv16    & 273M & 154M & 12/768/256 & 0/16  & 64+64 / 64 & 0.39M & $6.0 \times 10^{-4}$ \\
\midrule
EG-MLA-Base-emb64   & 155M & 38M & 12/768/64 & 0/64 & 64+64 / 64 & 0.39M & $6.0 \times 10^{-4}$ \\
EG-MLA-Base-emb128  & 195M & 77M & 12/768/128 & 0/64 & 64+64 / 64 & 0.39M & $6.0 \times 10^{-4}$ \\
EG-MLA-Base-emb256  & 274M & 154M & 12/768/256 & 0/64  & 64+64 / 64 & 0.39M & $6.0 \times 10^{-4}$ \\
EG-MLA-Base-emb512  & 433M & 309M & 12/768/512 & 0/64  & 64+64 / 64 & 0.39M & $6.0 \times 10^{-4}$ \\
EG-MLA-Base-emb1024 & 752M & 618M & 12/768/1024 & 0/64  & 64+64 / 64 & 0.39M & $6.0 \times 10^{-4}$ \\
EG-MLA-Base-emb2048 & 1389M & 1236M & 12/768/2048 & 0/64  & 64+64 / 64 & 0.39M & $6.0 \times 10^{-4}$ \\
\midrule
EG-MLA-XLarge-kv128emb512  & 1800M & 618M & 24/2048/512 & 0/128  & 128+64 / 128 & 0.39M & $2.0 \times 10^{-4}$ \\
EG-MLA-Large-kv128emb512  & 274M & 154M & 12/2048/512 & 0/128  & 128+64 / 128 & 0.39M & $2.5 \times 10^{-4}$ \\
EG-MLA-Large-kv256emb512  & 274M & 154M & 12/768/512 & 0/256  & 64+64 / 64 & 0.39M & $2.5 \times 10^{-4}$ \\

\bottomrule
\end{tabular}%
}
\caption{Sizes, architectures, and learning hyper-parameters (batch size in tokens and learning rate) of the models which we trained. All models were trained for a total of 50 billion tokens.}
\end{table}

\section{B. Why Reducing KV Cache Size Matters}
\subsection{Why is it so important to reduce the size of the KV cache?}

It is well known that inference for large language models (LLMs) is typically performed on GPUs, where memory is a scarce and valuable resource. On a single GPU, memory is primarily consumed by two components: (1) model parameters and activation values required for forward computation, which are relatively fixed once the model size is determined; and (2) the \textbf{KV cache}, which stores intermediate key and value tensors for autoregressive decoding. Unlike the model parameters, the KV cache grows dynamically with the input context length and can dominate total memory usage as the sequence becomes longer. In extreme cases, the KV cache alone can exceed the memory capacity of a single GPU or even an entire 8-GPU server.

A fundamental principle of LLM deployment is: \textit{if the model can run on a single GPU, do not use multiple GPUs; if it can run on a single server, do not distribute it across machines.} This is due to the hardware hierarchy: \textbf{intra-GPU communication bandwidth} is much higher than \textbf{inter-GPU}, which in turn is much higher than \textbf{inter-node} bandwidth. As a result, the more devices involved in inference, the more likely it is to suffer from communication bottlenecks—a classic manifestation of the \textit{bottleneck effect}. In fact, even the H100 GPU, with up to 3~TB/s of internal SRAM-to-HBM bandwidth, can still experience bandwidth constraints for short-context inference. These issues are exacerbated when spanning multiple GPUs or nodes with slower interconnects.

Therefore, \textbf{reducing KV cache size} enables inference over longer contexts using fewer devices, or allows larger batch sizes under the same memory budget. This results in either faster inference or higher overall throughput. Ultimately, all of these benefits translate to \textbf{lower inference cost}, which is a key concern in the practical deployment of LLMs.

\section{C. Detailed Architecture and Full Formulas}
\label{app:full_arch}

\begin{figure*}[t]
    \centering
    \includegraphics[width=0.95\textwidth]{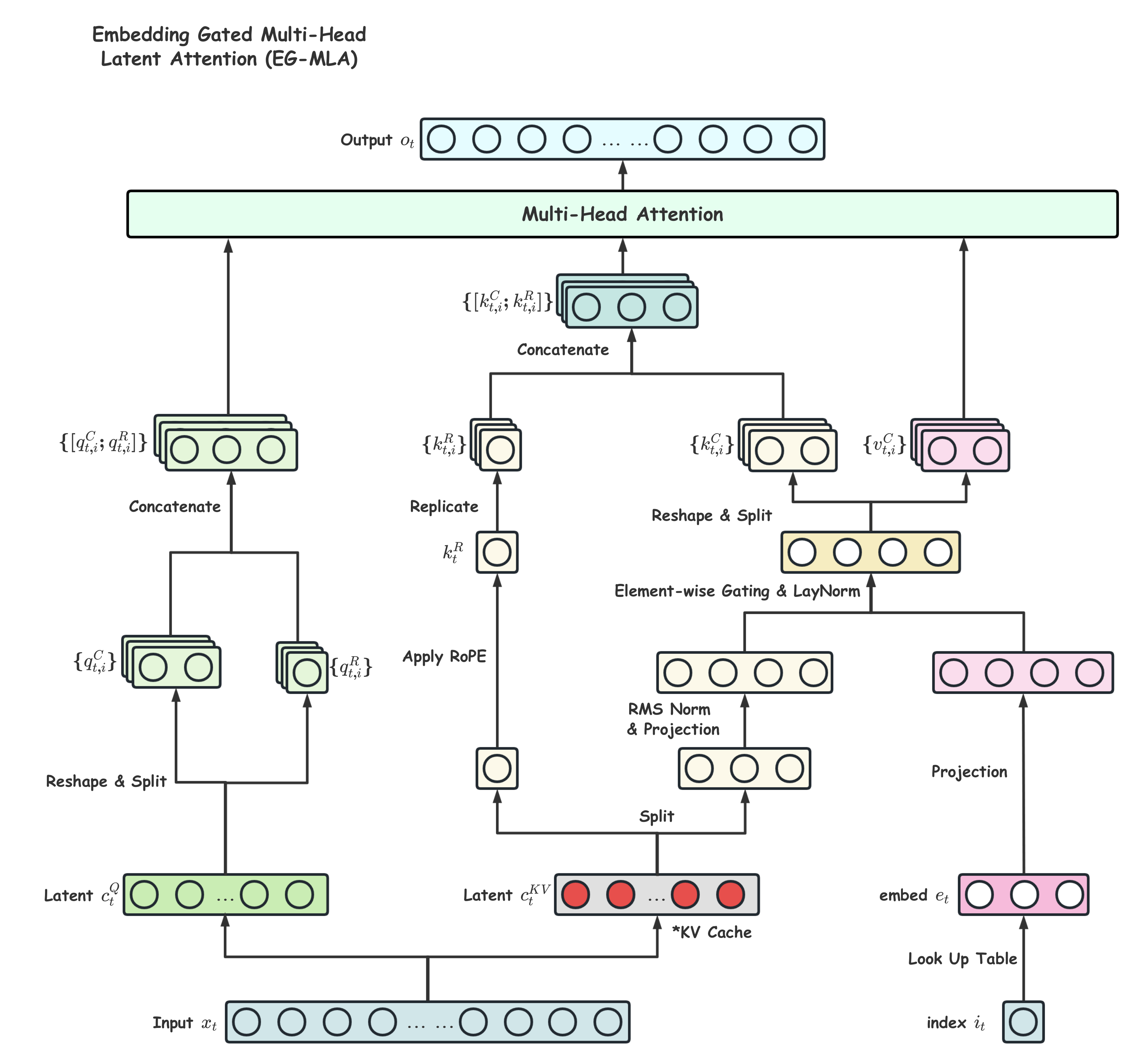}
    \caption{Detailed Architecture of the Embedding Gated Multi-Head Latent Attention (EG-MLA).}
    \label{fig:full_eg-mla}
\end{figure*}

Detailed Architecture is shown in Figure \ref{fig:full_eg-mla}.

The full computation pipeline begins by projecting the input into a low-rank latent space, followed by an up-projection to the original dimensionality. Embedding gating is then applied to produce the key and value representations for attention. These representations are subsequently split, with Partial RoPE individually applied to the keys and values, before being fed into a multi-head attention (MHA) module for final computation. Full formulas are shown as follows:

\setcounter{equation}{0}
\begin{align}
    \mathbf{c}_{t}^{KV} &= W^{DKV} \mathbf{x}_{t}, \\
    \mathbf{kv}_{t}^{C} &= W^{UKV} \mathbf{c}_{t}^{KV},\\
    \mathbf{e}_{t} &= \text{Emb}(i_t), \\
    \mathbf{g}_{t} &= W^{\text{UE}} \mathbf{e}_{t}, \\
    \widetilde{\mathbf{kv}}_{t}^{C} &= \text{LN} \left( \mathbf{kv}_{t}^{C} \odot \mathbf{g}_{t} \right),\\
    \mathbf{k}_{t}^{C},\mathbf{v}_{t}^{C} &= split(\widetilde{\mathbf{kv}}_{t}^{C})\\
    \mathbf{c}_{t}^{Q} &= W^{DQ} \mathbf{x}_{t}, \\
    \mathbf{q}_{t}^{C} &= W^{UQ} \mathbf{c}_{t}^{Q},\\
    \mathbf{q}_{t}^{R} &= \operatorname{RoPE}({W^{QR}} \mathbf{c}_{t}^{Q}), \\
    \mathbf{k}_{t}^{R} &= \operatorname{RoPE}({W^{KR}} \mathbf{x}_{t}), \\
    \mathbf{q}_{t, i} &= [\mathbf{q}_{t, i}^{C}; \mathbf{q}_{t, i}^{R}], \\
    \mathbf{k}_{t, i} &= [\mathbf{k}_{t, i}^{C}; \mathbf{k}_{t}^{R}], \\
    \mathbf{u}_{t, i} &= \sum_{j=1}^{t} \operatorname{Softmax}_j(\frac{\mathbf{q}_{t, i}^T \mathbf{k}_{j, i}}{\sqrt{d_{h} + d_{h}^{R}}}) \mathbf{v}_{j, i}^{C}, \\ 
    \mathbf{o}_{t} &= W^{O} [\mathbf{u}_{t, 1};\mathbf{u}_{t, 2};...;\mathbf{u}_{t, n_{h}}],
\end{align}

\section{D. More Experimental Results}
\label{appendix: More-Experimental-Results}
Here we present more experimental data, covering additional dimensions, a broader range of metrics, and more test datasets.

\subsection{Scaling Embedding Size}

\begin{table}[h]
\centering
\begin{tabular}{lcc|ccc|c}
\toprule
\textbf{Model Name} & \textbf{Embed Size} & \textbf{Tokens} & \textbf{ClimbMix-val} & \textbf{wiki} & \textbf{lambada} & \textbf{Avg.} \\ 
\midrule
EG-MLA-emb64 & 38M & 50B & 15.47 & 15.49 & 50.38 & 27.78 \\ 
EG-MLA-emb128 & 77M & 50B & 15.04 & 15.04 & 45.11 & 25.06 \\ 
EG-MLA-emb256 & 154M & 50B & 15.32 & 14.58 & 38.94 & 22.95 \\ 
EG-MLA-emb512 & 309M & 50B & 14.61 & 14.02 & 32.89 & 20.51 \\ 
EG-MLA-emb1024 & 618M & 50B & 13.91 & 13.72 & 35.41 & 21.01 \\ 
EG-MLA-emb2048 & 1236M & 50B & 13.81 & 13.62 & 34.25 & 20.56 \\ 
\bottomrule
\end{tabular}
\caption{Perplexity results of EG-MLA models with scaling embedding size. The KV cache is fixed at 64.}
\label{tab:fix-kv-50B-acc}
\end{table}

\subsection{Scaling KV Cache}

\begin{table}[h]
\centering
\begin{tabular}{lcc|ccc|c}
\toprule
\textbf{Model Name} & \textbf{Size} & \textbf{Tokens} & \textbf{ClimbMix-val} & \textbf{wiki} & \textbf{lambada} & \textbf{Avg.} \\
\midrule
MLA-kv256 & 120M & 50B & 15.45 & 15.63 & 49.88 & 26.99 \\
EG-MLA-kv256 & 125M & 50B & 14.32 & 14.26 & 37.12 & 21.90 \\
EG-MLA-kv128 & 122M & 50B & 14.49 & 14.34 & 35.73 & 21.52 \\
EG-MLA-kv64 & 120M & 50B & 15.32 & 14.58 & 38.94 & 22.95 \\
EG-MLA-kv16 & 119M & 50B & 15.13 & 15.08 & 43.76 & 24.66 \\
\bottomrule
\end{tabular}
\caption{Perplexity results of EG-MLA models across 12 tasks with scaling KV cache. The embedding size is fixed at 256. "Size" denotes the number of activated parameters.}
\label{tab:fix-emb-50B-ppl}
\end{table}

\section{E. PyTorch Code}
\label{app:code}

We provide the PyTorch code here to facilitate readers' understanding of our method.

\begin{lstlisting}[language=Python, style=mypython, escapechar=|, caption=Implementation of Embedding Gated MLA in pseudo PyTorch style., label={lst:pytorch}]
class EmbeddingGatedMLA(nn.Module):
    def __init__(self, config):
        super().__init__()
        self.n_embd = config.n_embd
        self.n_head = config.n_head
        self.qk_nope_head_dim = config.qk_nope_head_dim
        self.qk_rope_head_dim = config.qk_rope_head_dim
        self.qk_head_dim = config.qk_nope_head_dim + config.qk_rope_head_dim
        self.v_head_dim = config.v_head_dim
        self.q_lora_rank = config.q_lora_rank
        self.kv_lora_rank = config.kv_lora_rank
        assert config.n_embd % config.n_head == 0

        if self.q_lora_rank == 0:
            self.wq = nn.Linear(config.n_embd, config.n_head * self.qk_head_dim, bias=False)
        else:
            self.wq_a = nn.Linear(config.n_embd, config.q_lora_rank, bias=False)
            self.q_norm = nn.RMSNorm(config.q_lora_rank)
            self.wq_b = nn.Linear(config.q_lora_rank, config.n_head * self.qk_head_dim, bias=False)

        self.wkv_a = nn.Linear(config.n_embd, config.kv_lora_rank + config.qk_rope_head_dim, bias=False)
        self.kv_norm = nn.RMSNorm(config.kv_lora_rank)
        self.wkv_b = nn.Linear(config.kv_lora_rank, config.n_head * (config.qk_nope_head_dim + config.v_head_dim), bias=False)
        self.kv_emb = nn.Embedding(config.vocab_size, config.kv_emb_dim)
        self.kv_up = nn.Linear(config.kv_emb_dim, config.n_head * (config.qk_nope_head_dim + config.v_head_dim), bias=False)
        self.ln_emb = nn.LayerNorm(config.n_head * (config.qk_nope_head_dim + config.v_head_dim))
        self.c_proj = nn.Linear(config.n_head * config.v_head_dim, config.n_embd, bias=False)

    def forward(self, x, idx, freqs_cis):
        B, T, _ = x.size()
        if self.q_lora_rank == 0:
            q = self.wq(x)
        else:
            q = self.wq_b(self.q_norm(self.wq_a(x)))

        q = q.view(B, T, self.n_head, self.qk_head_dim)
        q_nope, q_pe = torch.split(q, [self.qk_nope_head_dim, self.qk_rope_head_dim], dim=-1)
        q_pe = apply_rotary_emb(q_pe, freqs_cis)

        kv = self.wkv_a(x)
        kv, k_pe = torch.split(kv, [self.kv_lora_rank, self.qk_rope_head_dim], dim=-1)
        k_pe = apply_rotary_emb(k_pe.unsqueeze(2), freqs_cis)

        q = torch.cat([q_nope, q_pe], dim=-1)
        kv = self.wkv_b(self.kv_norm(kv)) 
        kv = self.ln_emb(kv * self.kv_up(self.kv_emb(idx)))
        kv = kv.view(B, T, self.n_head, self.qk_nope_head_dim + self.v_head_dim)
        k_nope, v = torch.split(kv, [self.qk_nope_head_dim, self.v_head_dim], dim=-1)
        k = torch.cat([k_nope, k_pe.expand(-1, -1, self.n_head, -1)], dim=-1)

        k = k.view(B, T, self.n_head, -1).transpose(1, 2)
        q = q.view(B, T, self.n_head, -1).transpose(1, 2)
        v = v.view(B, T, self.n_head, -1).transpose(1, 2)

        y = F.scaled_dot_product_attention(q, k, v, attn_mask=None, is_causal=True)
        y = y.transpose(1, 2).contiguous().view(B, T, -1)

        y = self.c_proj(y)
        return y
\end{lstlisting}

\section{F. Comparison Between Attention Mechanisms}
\subsection{Comparison of Actual KV Cache Among MHA, MLA, and EG-MLA}
To ensure a fair evaluation of attention mechanism efficiency, all comparisons are conducted across models with comparable overall sizes and core architectural hyperparameters—namely, the same number of layers, model hidden dimension, and attention head configuration.

Table~\ref{tab:kv_cache_saving} highlights the dramatic reduction in KV cache size achieved by EG-MLA compared to standard Multi-Head Attention (MHA). 
While MHA stores separate key and value vectors for each attention head, resulting in 18.43K elements per token, EG-MLA reduces this number by up to \textbf{94.8\%}, requiring as few as 0.96K elements per token in the most compressed configuration.

Among the EG-MLA variants, EG-MLA-kv64 offers a particularly favorable trade-off between performance and memory efficiency. 
It reduces the KV cache per token to only \textbf{1.54K} elements, which constitutes a \textbf{91.6\% reduction} compared to standard Multi-Head Attention (MHA), and a \textbf{59.9\% reduction} relative to MLA. 
This substantial saving is achieved without altering the number of attention heads or compromising activated model parameters.

\begin{table}[ht]
\centering
\resizebox{\textwidth}{!}{%
\begin{tabular}{lccccccc}
\toprule
\textbf{Model Name} & 
\textbf{Total Params} & 
\textbf{Activated Params} & 
\textbf{Embed Params} & 
\makecell[c]{\textbf{n\_layer} / \\ \textbf{n\_emb} / \\ \textbf{kv\_emb\_dim}} & 
\makecell[c]{\textbf{q\_lora\_rank} / \\ \textbf{kv\_lora\_rank}} & 
\makecell[c]{\textbf{qk\_head\_dim} / \\ \textbf{v\_head\_dim}} & 
\makecell[c]{\textbf{KV Cache / Token} \\ \textbf{(\#Element)}} \\
\midrule
MHA-Base      & 124M & 124M & 0M & 12/768/- & - & 64 & 18.43K \\
MLA-Base      & 120M & 120M & 0M & 12/768/256 & 0/256 & 64+64 / 64 & 3.84K \\
\midrule
EG-MLA-Base-kv256   & 279M & 125M & 154M & 12/768/256 & 0/256 & 64+64 / 64 & 3.84K \\
EG-MLA-Base-kv128   & 276M & 122M & 154M & 12/768/256 & 0/128 & 64+64 / 64 & 2.30K \\
EG-MLA-Base-kv64    & 274M & 120M & 154M & 12/768/256 & 0/64  & 64+64 / 64 & 1.54K \\
EG-MLA-Base-kv16    & 273M & 119M & 154M & 12/768/256 & 0/16  & 64+64 / 64 & 0.96K \\
\bottomrule
\end{tabular}%
}
\caption{
Comparison of actual KV cache usage among Multi-Head Attention (MHA), Multi-Head Latent Attention (MLA), and our proposed Embedding Gated MLA (EG-MLA). 
EG-MLA achieves comparable or stronger performance while significantly reducing the KV cache per token, thanks to its compressed latent representation and token-specific gating.
}
\label{tab:kv_cache_saving}
\end{table}

\section{G. More Scaling Experiment}
\label{appendix: More-Scaling-Experiment}
\begin{figure}[H]
    \centering
    \includegraphics[width=\linewidth]{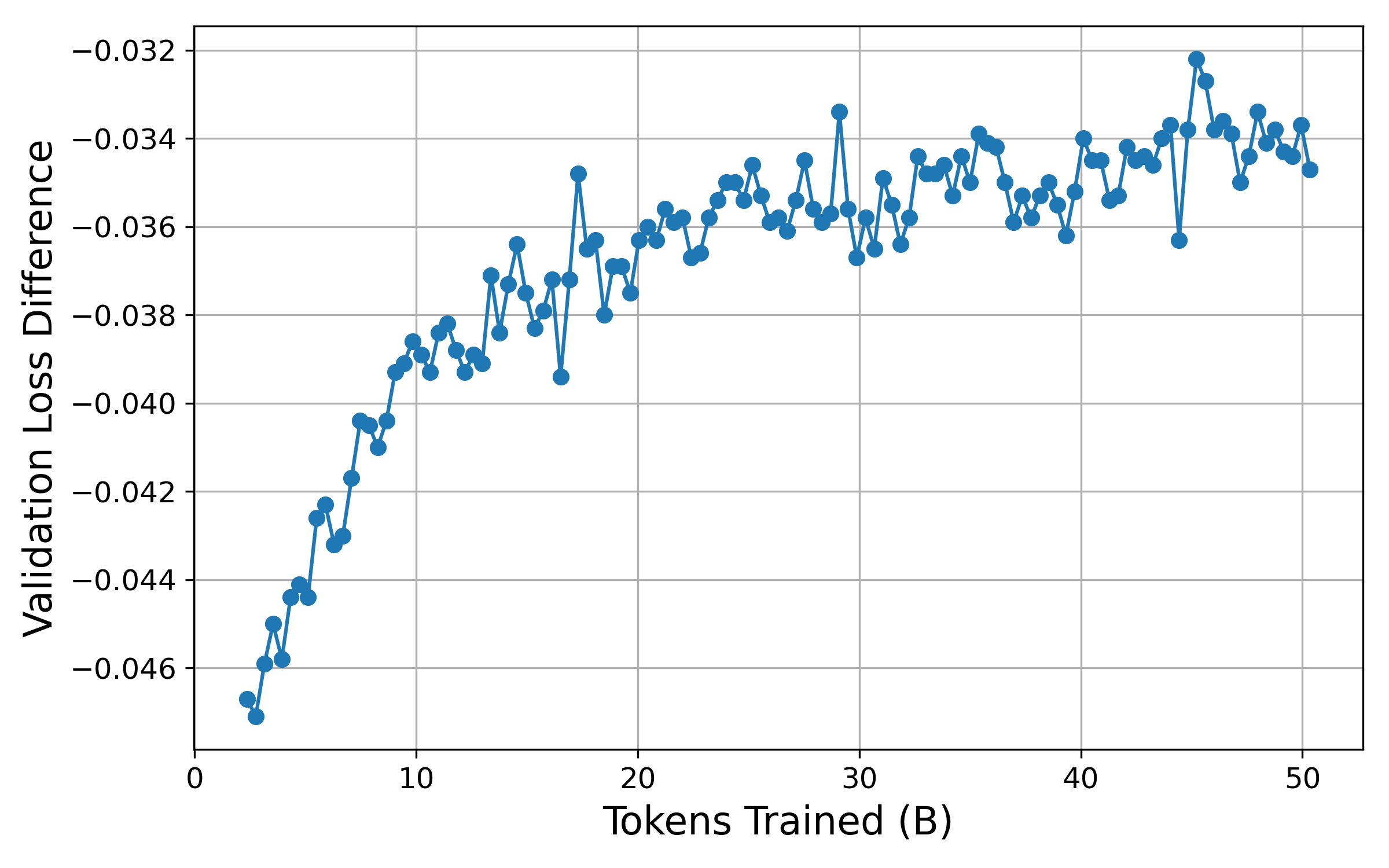}
    \caption{Difference in validation loss between EG-MLA-1.2B and MLA-1.2B models as a function of tokens trained.}
    \label{fig:1B2_val_loss_diff}
\end{figure}

Figure~\ref{fig:1B2_val_loss_diff} shows the difference in validation loss between the EG-MLA-1.2B model and the baseline MLA-1.2B model over the course of training. The x-axis represents the number of tokens processed (in billions), while the y-axis indicates the validation loss difference. A negative value implies that EG-MLA-1.2B performs better (i.e., lower validation loss) than the baseline at that point in training. The trend highlights how the enhancements in EG-MLA-1.2B translate into improved generalization throughout the training trajectory.

\begin{table*}[h]
\resizebox{\linewidth}{!}{%
\begin{tabular}{lcc|cccccccccccc|c}
\midrule
\textbf{Model Name} &
  \textbf{Model Size} &
  \textbf{Embed Size} &
  \textbf{piqa} &
  \textbf{arc-c} &
  \textbf{arc-e} &
  \textbf{hella} &
  \textbf{winog} &
  \textbf{siqa} &
  \textbf{mmlu} &
  \textbf{obqa} &
  \textbf{boolq} &
  \textbf{race} &
  \textbf{truth} &
  \textbf{lmb.o} &
  \textbf{Avg.} \\ 
\midrule
EG-MLA-Large-ctx2048-kv128emb512 & 951M & 309M & 75.35 & 37.20 & 70.33 & 42.50 & 53.83 & 90.80 & 27.22 & 26.80 & 57.55 & 33.01 & 29.61 & 39.72 & 48.66 \\ 
MLA-Large-ctx2048-kv512 & 645M & 0M & 74.54 & 37.88 & 69.70 & 42.16 & 55.01 & 90.40 & 27.34 & 27.20 & 62.97 & 34.83 & 31.12 & 41.24 & 49.53 \\ 
EGA-XLarge-ctx4096-kv128emb512 & 1800M & 618M & 75.41 & 38.74 & 72.85 & 44.88 & 56.83 & 91.00 & 26.92 & 27.80 & 59.45 & 33.97 & 31.27 & 40.25 & 49.95 \\ 
MLA-XLarge-ctx4096-kv512 & 1188M & 0M & 75.52 & 38.82 & 73.02 & 44.15 & 56.83 & 92.20 & 28.42 & 27.20 & 53.15 & 34.26 & 33.76 & 38.02 & 49.61 \\ 
EG-MLA-Large-ctx2048-kv256emb512 & 961M & 309M & 74.92 & 37.12 & 71.55 & 42.79 & 54.38 & 90.70 & 25.93 & 23.80 & 58.96 & 33.21 & 29.24 & 41.98 & 48.71 \\ 
\bottomrule
\end{tabular}%
}
\caption{Accuracy results of more models across 12 benchmark tasks. \emph{kv} refers to KV cache
dimension and \emph{emb} refers to embedding dimension.All models were trained for a total of 50 billion tokens.}
\end{table*}

\begin{table}[h]
\centering
\begin{tabular}{lcc|ccc|c}
\toprule
\textbf{Model Name} & \textbf{Model Size} & \textbf{Embed Size} & \textbf{ClimbMix-val} & \textbf{wiki} & \textbf{lambada} & \textbf{Avg.} \\
\midrule
EG-MLA-Large-ctx2048-kv128emb512 & 951M & 309M & 10.79 & 6.17 & 19.57 & 12.18 \\
MLA-Large-ctx2048-kv512 & 645M & 0M & 11.02 & 6.31 & 18.95 & 12.09 \\
EGA-XLarge-ctx4096-kv128emb512 & 1800M & 618M & 10.35 & 2.94 & 19.01 & 10.77 \\
MLA-XLarge-ctx4096-kv512 & 1188M & 0M & 10.71 & 3.00 & 22.99 & 12.23 \\
EG-MLA-Large-ctx2048-kv256emb512 & 961M & 309M & 10.58 & 6.14 & 17.95 & 11.56 \\
\bottomrule
\end{tabular}
\caption{Perplexity results of more models across 12 benchmark tasks. \emph{kv} refers to KV cache
dimension and \emph{emb} refers to embedding dimension.All models were trained for a total of 50 billion tokens.}
\end{table}

\section{H. Compression through Layer Grouping}
\label{appendix:CompressionThroughLayerGrouping}
To push the optimization frontier and further curtail computational and memory costs, we explore the reuse of both the KV-projection parameters and their intermediate activations across transformer layers. Specifically, we introduce a hyper-parameter $layer group size(lgz)$ that dictates how many consecutive transformer layers share a single parameter set rather than maintaining independent parameters for each layer. During the forward pass, all layers within the same group directly reuse the shared parameters and cached intermediate results, eliminating redundant computation and improving training and inference efficiency.

Empirical results show that when $lgz$ = 2, the performance of model remains comparable to the baseline. However, once $lgz$ exceeds 4, a significant performance degradation is observed. Therefore, $lgz$ = 4 represents the compression limit under current settings, and we recommend $lgz$ = 3 as a balanced trade-off between efficiency gains and accuracy loss.

\begin{table*}[h]
\resizebox{\linewidth}{!}{%
\begin{tabular}{l|cccccccccccc|cc}
\midrule
\textbf{Model Name} &
  \textbf{piqa} &
  \textbf{arc-c} &
  \textbf{arc-e} &
  \textbf{hella} &
  \textbf{winog} &
  \textbf{siqa} &
  \textbf{mmlu} &
  \textbf{obqa} &
  \textbf{boolq} &
  \textbf{race} &
  \textbf{truth} &
  \textbf{lmb.o} &
  \textbf{Avg.} &
  \textbf{Diff (\%)} \\ 
\midrule
MLA-Large-ctx2048-kv512 & 74.54 & 37.88 & 69.70 & 42.16 & 55.01 & 90.40 & 27.34 & 27.20 & 62.97 & 34.83 & 31.12 & 41.24 & 49.53 & --- \\
EG-MLA-Large-ctx2048-kv256emb512 & 74.92 & 37.12 & 71.55 & 42.79 & 54.38 & 90.70 & 25.93 & 23.80 & 58.96 & 33.21 & 29.24 & 41.98 & 48.71 & --- \\
EG-MLA-Large-ctx2048-kv256emb512lgz2 & 74.76 & 37.03 & 69.57 & 42.56 & 52.33 & 90.50 & 25.65 & 27.80 & 59.79 & 31.87 & 30.59 & 39.90 & 48.53 & -0.37 \\
EG-MLA-Large-ctx2048-kv256emb512lgz4 & 75.03 & 36.18 & 69.32 & 41.61 & 52.25 & 91.60 & 27.70 & 26.40 & 55.69 & 32.25 & 29.27 & 40.50 & 48.15 & -1.15 \\
EG-MLA-Large-ctx2048-kv256emb512lgz6 & 74.48 & 36.35 & 70.58 & 40.69 & 52.80 & 90.10 & 26.98 & 24.60 & 58.10 & 31.20 & 29.20 & 38.48 & 47.80 & -1.87 \\
\bottomrule
\end{tabular}%
}
\caption{Accuracy results across 12 benchmark tasks under different layer group sizes (lgz). Diff is the relative change (\%) compared to EG-MLA-Large-ctx2048-kv256emb512.}
\end{table*}

\begin{table}[h]
\centering
\begin{tabular}{l|ccc|cc}
\toprule
\textbf{Model Name} & \textbf{ClimbMix-val} & \textbf{wiki} & \textbf{lambada} & \textbf{Avg.} & \textbf{Diff (\%)} \\
\midrule
MLA-Large-ctx2048-kv512 & 11.02 & 6.31 & 18.95 & 12.09 & --- \\
EG-MLA-Large-ctx2048-kv256emb512 & 10.58 & 6.14 & 17.95 & 11.56 & --- \\
EG-MLA-Large-ctx2048-kv256emb512lgz2 & 10.78 & 6.23 & 18.69 & 11.90 & -2.94 \\
EG-MLA-Large-ctx2048-kv256emb512lgz4 & 11.05 & 6.38 & 20.19 & 12.54 & -8.47 \\
EG-MLA-Large-ctx2048-kv256emb512lgz6 & 11.27 & 6.51 & 22.77 & 13.52 & -16.98 \\
\bottomrule
\end{tabular}
\caption{Perplexity results across 12 benchmark tasks under different layer group sizes (lgz). Diff is the relative change (\%) compared to EG-MLA-Large-ctx2048-kv256emb512.}
\end{table}